\documentclass[10pt,twocolumn,letterpaper]{article}

\usepackage[cvprfinal]{cvpr2025template/cvpr} 

\definecolor{cvprblue}{rgb}{0.21,0.49,0.74}
\usepackage[pagebackref,breaklinks,colorlinks,allcolors=cvprblue]{hyperref}

\def\paperID{13796} %
\def\confName{CVPR}
\def\confYear{2025}

\usepackage[utf8]{inputenc} %
\usepackage[T1]{fontenc}    %
\usepackage{hyperref}       %
\usepackage{url}            %
\usepackage{booktabs}       %
\usepackage{amsfonts}       %
\usepackage{nicefrac}       %
\usepackage{microtype}      %
\usepackage{xcolor}         %

\newcommand{\veryshortarrow}[1][2pt]{\mathrel{%
   \hbox{\rule[\dimexpr\fontdimen22\textfont2-.2pt\relax]{#1}{.4pt}}%
   \mkern-4mu\hbox{\usefont{U}{lasy}{m}{n}\symbol{41}}}}

\makeatletter

\setbox0\hbox{$\xdef\scriptratio{\strip@pt\dimexpr
    \numexpr(\sf@size*65536)/\f@size sp}$}

\newcommand{\scriptveryshortarrow}[1][3pt]{{%
    \hbox{\rule[\scriptratio\dimexpr\fontdimen22\textfont2-.2pt\relax]
               {\scriptratio\dimexpr#1\relax}{\scriptratio\dimexpr.4pt\relax}}%
   \mkern-4mu\hbox{\let\f@size\sf@size\usefont{U}{lasy}{m}{n}\symbol{41}}}}

\makeatother

\newcommand{\mypar}[1]{\vspace{-2.5mm}\paragraph{#1}}

\newcommand{\xhdr}[1]{\vspace{5pt}\noindent\textbf{#1}}

\newcommand{\cmark}{\ding{51}}%

\definecolor{graycolor}{gray}{.9}

\usepackage{amsmath,amsfonts,bm}

\def\eqref#1{equation~\ref{#1}}

\def\1{\bm{1}}

\DeclareMathAlphabet{\mathsfit}{\encodingdefault}{\sfdefault}{m}{sl}
\SetMathAlphabet{\mathsfit}{bold}{\encodingdefault}{\sfdefault}{bx}{n}

\def\gL{{\mathcal{L}}}

\def\sR{{\mathbb{R}}}

\newcommand{\E}{\mathbb{E}}

\usepackage{graphicx}
\usepackage{amsmath}
\usepackage{colortbl}
\usepackage{cuted}
\usepackage{multirow}
\usepackage{arydshln}
\usepackage{comment}
\usepackage{enumitem}
\usepackage{tabularx}
\usepackage{float}
\usepackage{multicol}

\usepackage{pifont}%

\usepackage{wrapfig,lipsum,booktabs}

\newcommand{\redcolor}[1]{\textcolor{red}{#1}}

\title{Self-Supervised Spatial Correspondence Across Modalities}
\author{Ayush Shrivastava \qquad 
Andrew Owens\\
University of Michigan\\
\vspace{1em}{\small \url{https://ayshrv.com/cmrw}}
}

\begin{document}

\maketitle

\vspace{-1mm}
\begin{strip}
    \centering
    \vspace{-18mm}
    \includegraphics[width=\linewidth]{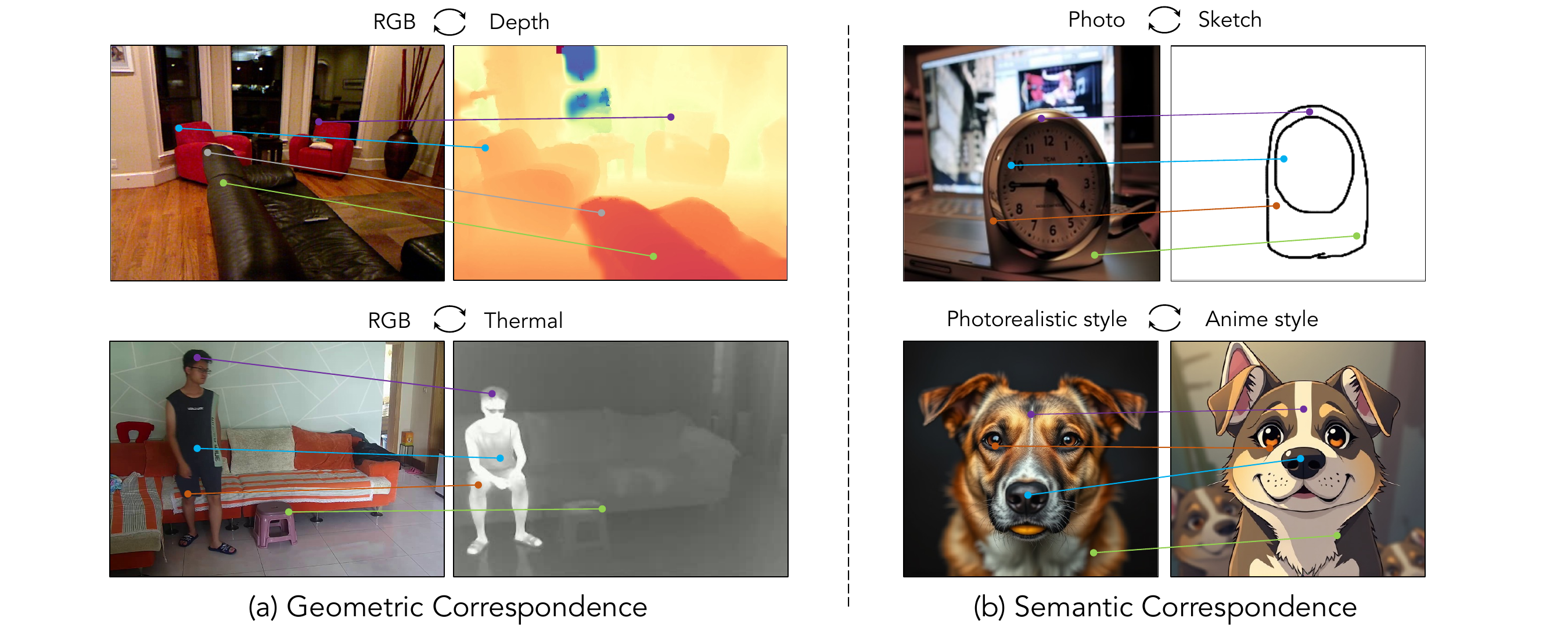}\vspace{-3mm}%

    \captionof{figure}{{\bf Finding spatial correspondences across modalities. } We present a method for cross-modal matching, trained entirely through self-supervision using a simple formulation based on contrastive random walks~\cite{jabri2020space}. (a) Given two images taken by different visual modalities and at different positions and times, we predict the pairs of image patches that physically correspond to the same points. (b) We also apply our method to semantic matching tasks, using a visual encoder initialized with pretrained DINOv2~\cite{oquab2023dinov2} weights and fine-tuned during training. These include tasks such as photo-sketch alignment~\cite{lu2023learning} and style-based matching between images of different styles generated with a text-to-image model~\cite{flux2024}.}
    \label{fig:teaser}
    \vspace{-2mm}
\end{strip}

\begin{abstract}
\vspace{-6mm}

We present a method for finding cross-modal space-time correspondences.  Given two images from different visual modalities, such as an RGB image and a depth map, our model identifies which pairs of pixels correspond to the same physical points in the scene. To solve this problem, we extend the contrastive random walk framework to simultaneously learn cycle-consistent feature representations for both cross-modal and intra-modal matching. The resulting model is simple and has no explicit photo-consistency assumptions. It can be trained entirely using unlabeled data, without the need for any spatially aligned multimodal image pairs. We evaluate our method on both geometric and semantic correspondence tasks. For geometric matching, we consider challenging tasks such as RGB-to-depth and RGB-to-thermal matching (and vice versa); for semantic matching, we evaluate on photo-sketch and cross-style image alignment. Our method achieves strong performance across all benchmarks.

\end{abstract}

\section{Introduction}

Cameras take a multitude of different forms. While RGB, thermal, and depth cameras each record images, what is stored within each pixel differs drastically.
Consequently, when a scene is captured using cameras that are based on different sensory modalities, it is difficult to determine which pixels within them correspond to the same physical points.  These cross-modal pixel correspondences have a number of applications, such as cross-modal registration, 3D reconstruction, and multimodal data fusion~\cite{arar2020unsupervised,  dou2024tactileaugmented, li20232d3d,wang2023freereg}. %
Recent work on multimodal learning has provided effective ways of learning cross-modal correspondences through self-supervision, yet these methods largely rely on having paired data from multiple sensors, which is not available for pixel-level correspondence.

Traditional self-supervised methods for pixel correspondences often make strong assumptions about the visual appearance of  pixels that should be matched together, such as by assuming that matching image patches should be photoconsistent~\cite{jonschkowski2020matters,jason2016back,stone2021smurf,janai2018unsupervised,zou2018df}, or that cross-modal translation methods can estimate one modality from another as part of the matching process~\cite{arar2020unsupervised}. However, these assumptions are frequently violated in multimodal data, making it challenging to use them as general-purpose multimodal matching methods. Depth images and RGB images, for example, may look alike to a human, but their intensity values record information that is so different---depth {\em vs.} lightness---that they cannot be matched using local information. And cross-modal prediction requires being able to solve a notoriously difficult problem---monocular depth estimation---without explicit paired data.

We take inspiration from recent advances in self-supervised tracking~\cite{wang2019learning,jabri2020space,bian2022learning,li2019joint} that learn correspondences between video frames through {\em cycle consistency}.  These methods assume that the content within a scene persists between frames, and that there should thus be a one-to-one correspondence between the pixels in them.  Our approach extends the {contrastive random walk}~\cite{jabri2020space} to the {\em cross-modal} pixel correspondence problem.  We create a directed graph in which nodes correspond to image patches in each modality, and where edges connect patches across modalities.  We train a network, based on the recently proposed global matching transformer~\cite{shrivastava2024gmrw,xu2022gmflow}, to assign transition probabilities to pairs of patches for a random walk.  A random walker uses these transition probabilities to step through the graph, moving from patches in one modality to another, and then back.  To train the model, we maximize the walker's return probability, thus encouraging matches to be cycle-consistent.  

A key challenge in our setting is that, unlike many correspondence tasks, the patches we match are visually very different.  We therefore incorporate {\em intra-modal}  random walks between images of the same modality, such that the model must learn a representation that simultaneously allows it to match intra- and inter-modally. To get these image pairs, we apply data augmentation to input images, simulating the challenges of matching frames of a video. We also show that, in contrast to image-based matching~\cite{shrivastava2024gmrw}, spatial smoothness priors are important for obtaining strong performance. 
Moreover, in contrast to other cross-modal correspondence methods~\cite{arar2020unsupervised,jonschkowski2020matters,stone2021smurf}, our method does not rely on hand-crafted measures of photoconsistency.
Instead, it learns visual similarity through cycle-consistent representation learning, defined by the learned embedding space, thereby making it possible to apply it to different pairs of modalities without any modifications.

We evaluate our model on both geometric and semantic cross-modal correspondence tasks. For geometric matching, we focus on two challenging tasks: RGB-to-depth and RGB-to-thermal matching (Figure~\ref{fig:teaser}a). In both cases, we match images captured at different times and positions, with different sensors. For semantic matching (Figure~\ref{fig:teaser}b), we use a visual encoder initialized with pretrained DINOv2~\cite{oquab2023dinov2} weights and fine-tune it during training. We evaluate on photo-to-sketch alignment, where our method performs competitively with approaches specifically tailored to this task, and on a new cross-style image matching task, using images generated in different styles from the same diffusion model~\cite{flux2024}. Our work makes several contributions:
\begin{itemize}[leftmargin=10pt, topsep=1pt, noitemsep]
\item We show that space-time cross-modal pixel correspondence can be learned from unlabeled data through cycle consistency.
\item We extend the contrastive random walk framework for cross-modal pixel correspondence.
\item Through experiments, we show that our model successfully finds correspondences in geometric tasks such as RGB-to-depth and RGB-to-thermal, significantly outperforming prior methods. In semantic tasks like photo-to-sketch matching, our method is competitive with other self-supervised approaches. %
\item We propose benchmarks for evaluating RGB-to-depth, RGB-to-thermal, and cross-style image matching. For cross-style matching, we use generative models to synthesize images with similar content but different styles. %
\end{itemize}
\section{Related Work}
\begin{figure*}[t]
    \centering
    \includegraphics[width=\textwidth]{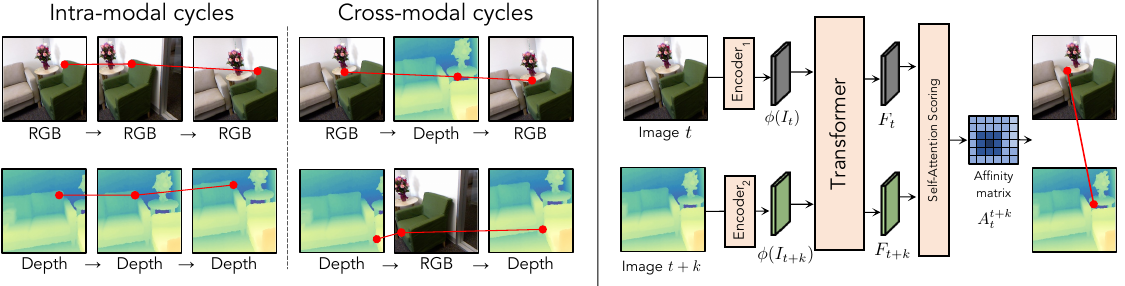}
    \vspace{-5mm}%
    \caption{\small{\small\bf Model Architecture.} We learn to find pixel-level correspondences between images that may differ in sensory modality, time, and scene position. Given images from two modalities (e.g., unpaired RGB and depth images from the same scene), we perform a contrastive random walk on a graph whose nodes come from patches within the two images using a global matching transformer architecture~\cite{shrivastava2024gmrw}. We simultaneously perform auxiliary intra-modal random walks within each modality's augmented crops of images to improve the model's ability to avoid local minima during optimization. Through this process, we learn to match in both directions (e.g., RGB-to-depth  and depth-to-RGB). }
    \label{fig:model_architecture}
    \vspace{-5mm}
\end{figure*}

\paragraph{Cycle consistency for correspondence.}
Zhou et al.~\cite{zhou2016learning} proposed learning correspondences across 2D views of 3D objects, by matching the viewpoints to a 3D model, then projecting back to other images.
In contrast, we focus on matching two images that differ in {\em modality}, {\em space}, and {\em time}, rather than just 3D viewpoint.
Later work extended cycle consistency to space-time tracking.
Wang et al.~\cite{wang2019learning} trained a model to track forward and backward in time, penalizing deviation from the original position, using a formulation based on the spatial transformer~\cite{jaderberg2015spatial}.
Jabri et al.\cite{jabri2020space} framed this as a random walk on a space-time graph, enabling self-supervised tracking, followed by improvements such as shortcut removal\cite{tang2021breaking} and smoothness priors~\cite{bian2022learning} for fine-grained correspondence. Recent works further integrated this framework with global matching transformers~\cite{xu2022gmflow,shrivastava2024gmrw} to address the tracking-any-point problem~\cite{doersch2022tap}.  

While these works focus on temporal matching, we extend contrastive random walks to cross-modal matching, introducing both intra- and inter-modal random walks. Although contrastive random walks have been explored in multimodal settings like image-audio alignment~\cite{hu2022mix} or audio-to-audio matching~\cite{chen2020simple}, these approaches operate at the patch or node level and do not aim for dense pixel-to-pixel correspondence, which is the focus of our work.

\mypar{Cross-modal pixel correspondence.}
Cross-modal correspondence has been explored through contrastive learning~\cite{tian2020contrastive,radford2021learning,he2020momentum}, but these approaches typically learn global or patch-level embeddings, not dense pixel-wise matches. Arar et al.\cite{arar2020unsupervised} proposed a method for RGB-infrared matching by jointly training a GAN for modality translation and a spatial transformation network. In contrast, our model does not require explicit cross-modal translation; it learns from cycle-consistency alone. Other methods such as\cite{lu2023learning} address photo-sketch alignment through a two-stage process involving cross-modal embeddings, whereas we learn fine-grained correspondence directly through self-supervision. Some recent work has explored matching across 2D-3D modalities (e.g., RGB and point clouds~\cite{li20232d3d,wang2023freereg}), but these approaches often rely on explicit geometric priors and operate on sparse data like LiDAR. Our method, by contrast, is designed for dense 2D image domains and does not rely on any hand-crafted similarity metric.

\mypar{Self-supervised correspondence.}
Unsupervised learning of correspondence has a long history in optical flow~\cite{jonschkowski2020matters,jason2016back,stone2021smurf,janai2018unsupervised,zou2018df}, often relying on photometric loss as a training signal. Other approaches use cues like color similarity~\cite{vondrick2018tracking,li2019joint,lai2020mast} or temporal coherence~\cite{xu2021rethinking,gordon2020watching} for self-supervised tracking. However, these techniques assume single-modality inputs and struggle to generalize to cross-modal settings. Simulated data has also been used for learning correspondence~\cite{sun2021autoflow,harley2022particle,greff2022kubric,doersch2022tap,doersch2023tapir}, but generating realistic, time-varying multimodal signals remains a significant challenge. In contrast, our method learns from real multimodal videos without requiring ground-truth annotations or simulation.

\section{Method}

Our goal is to learn pixel-wise correspondences between images from different modalities (e.g., RGB, thermal, depth, sketch, stylistically diverse images) solely through self-supervision. We build upon the contrastive random walk framework~\cite{jabri2020space} and extend it to handle multiple modalities (Figure~\ref{fig:model_architecture}). The recently proposed Global Matching Random Walk (GMRW) model~\cite{shrivastava2024gmrw} allows for dense correspondences between RGB images by enforcing cycle consistency via contrastive random walks over videos~\cite{xu2022gmflow,shrivastava2024gmrw}. We leverage this framework to train on unlabeled videos from different modalities, learning cycle-consistent tracking across pairs such as RGB-depth, RGB-thermal, photo-sketch, and cross-style images.

However, we find that directly optimizing the model for cross-modal cycle consistency leads to poor convergence: training plateaus at a suboptimal solution with high loss and non-semantic matches. We hypothesize that this difficulty arises due to the substantial domain gap between modalities. For instance, unlike space-time matching---where initial random features provide reasonable gradients---cross-modal inputs may align arbitrarily (e.g., bright regions in RGB could align with distant areas in depth due to similarly high intensity values). To mitigate this, we introduce auxiliary intra-modality random walk supervision, encouraging the model to learn representations effective for both intra- and cross-modality matching.

\subsection{Model architecture}
To address the challenge of cross-modal dense correspondence, we require an architecture capable of producing high-resolution matches while remaining modality-agnostic. We adopt the Global Matching Random Walk (GMRW) model~\cite{shrivastava2024gmrw}, which performs all-pairs pixel matching~\cite{teed2020raft,xu2022gmflow} and is trained using a cycle consistency objective via contrastive random walks. Crucially, it does not rely on modality-specific assumptions like photo-consistency. Each modality uses a dedicated visual encoder to produce input tokens, while a shared Transformer backbone performs the global matching across modalities.

\xhdr{Image Features.} For geometric tasks, we use a CNN-based encoder to extract visual features. For each image $I^{m}_{t}$ at time $t$ and modality $m\in\{\texttt{rgb}, \texttt{depth}, \texttt{thermal}\}$, we compute $d$-dimensional features $\phi(I^{m}_{t}) \in \sR^{\frac{H}{c}\times\frac{W}{c}\times d}$ with a downsampling factor $c=4$. Each modality has its own encoder, and we append 2D positional encodings to the features. For depth inputs, the single-channel image is replicated across three channels to match the RGB architecture.

For semantic tasks (e.g., photo-sketch, cross-style), we use DINOv2~\cite{oquab2023dinov2} as a shared visual encoder across modalities. We initialize it with pretrained weights and finetune it, leveraging the rich semantic priors learned during DINOv2 pretraining.

\xhdr{Cross-modal Matching Transformer.} Given a pair of images $I^{m_1}_{t}$ and $I^{m_2}_{t+k}$, we feed their extracted features into a shared Transformer consisting of six blocks of self-attention, cross-attention, and feed-forward layers. The final layer produces correlation features $F^{m_1}_{t}$ and $F^{m_2}_{t+k}$. This module supports both intra-modality ($m_1 = m_2$) and cross-modality ($m_1 \ne m_2$) matching.

\xhdr{Cross-modal Transition Matrix.} We compute the transition matrix from $F^{m_1}_{t}$ and $F^{m_2}_{t+k}$ as $A_{t, t+k}^{m_1, m_2} = \text{softmax}(F^{m_1}_{t} ({F^{m_{2}}_{t+k}})^{\top} / \tau)$, which represents the likelihood of a pixel in modality $m_1$ at time $t$ corresponding to a pixel in modality $m_2$ at time $t+k$. This matrix defines the transition probabilities for contrastive random walks (CRW) in both intra- and cross-modality matching. Following ~\cite{shrivastava2024gmrw, xu2022gmflow}, we normalize features using $\tau = \sqrt{d}$ instead of the L2 normalization used in earlier work~\cite{jabri2020space,bian2022learning,tang2021breaking}.

The expected change in pixel position is computed as: 
\begin{equation}
\mathbf{f}_{t, t+k}^{m_1, m_2} = \mathbb{E}_{A_{t,t+k}} [A_{t, t+k}^{m_1, m_2} D - D]
\end{equation}
where $\mathbf{f}_{t, t+k}$ is the predicted flow between modalities $m_1$ and $m_2$, $D$ is the constant pixel coordinate grid and  $A_{t, t+k}^{m_1, m_2}$ is the transition matrix from frame $t$ to $t+k$.

\subsection{Learning cross-modal correspondences}
\label{sec:losses}
\xhdr{Cross-modal cycle-consistency.} To learn cross-modal correspondences, we extend the cycle consistency objective from GMRW~\cite{shrivastava2024gmrw} to multi-modal data. Given a \textit{palindrome} sequence $\{I^{m_1}_{t}, I^{m_2}_{t+k}, I^{m_1}_{t}\}$, we treat it as a space-time graph and train our model to perform random walks across it. The model estimates two transition matrices: $A_{t, t+k}^{m_1, m_2}$ which maps pixels from modality $m_1$ at time $t$ to modality $m_2$ at time $t+k$, and $A_{t+k, t}^{m_2, m_1}$, which returns the walker back to modality $m_1$ at time $t$. We chain these two transitions to obtain a round-trip path from a point in $I^{m_1}_{t}$, through $I^{m_2}_{t+k}$, and back. To enforce cycle consistency, we maximize the probability that the walker returns to its original position. This is done using the label-warping loss from~\cite{shrivastava2024gmrw}:
\begin{equation}
\gL_{\mathtt{cross{\text -}crw}} = \gL_{\mathtt{CE}}( A_{t, t+k}^{m_1, m_2} A_{t+k, t}^{m_2, m_1}, T_f^b(I)),
\label{eq:inter_crw_loss}
\end{equation}
where $I$ is the identity matrix and $T_f^b(I)$ is the warped label target designed to prevent shortcut learning~\cite{shrivastava2024gmrw}, and $\gL_{\mathtt{CE}}$ is the cross-entropy loss.

\xhdr{Intra-modal cycle-consistency.} 
We further apply the same contrastive random walk objective within each modality to stabilize training.
For a palindrome $\{ I^{m_i}_{\mathtt{ori}},  I^{m_i}_{\mathtt{aug}}, I^{m_i}_{\mathtt{ori}} \}$, 
where $I^{m_i}_{\mathtt{aug}}$ is an augmented crop of $I^{m_i}_{\mathtt{ori}}$, the intra-modal cycle loss is:
\begin{equation}
\gL_{\mathtt{intra{\text -}crw}} = \sum_{i=1}^{2} \gL_{\mathtt{CE}}(A_{\mathtt{ori},\mathtt{aug}}^{m_i}\,  A_{\mathtt{aug},\mathtt{ori}}^{m_i}, T_f^b(I))
\label{eq:intra_crw_loss}
\end{equation}

\xhdr{Smoothness loss.} To encourage spatial coherence, we apply an edge-aware smoothness loss~\cite{jonschkowski2020matters} --- used in prior contrastive random walk work~\cite{bian2022learning,shrivastava2024gmrw}. It penalizes large second-order derivatives in flow, weighted by image gradients. We apply this loss only when the source image is from the RGB modality, where visual similarity is a reliable cue for perceptual grouping. The loss is defined as:
\begin{equation}
    \gL_{\mathtt{smooth}} = \E_p \sum_{d \in \{x, y\}} \text{exp} (-\lambda_c I_d(p)) |\frac{\partial^2 \textbf{f}_{s,t}(p)}{\partial d^2}|   
\end{equation}
where the $p$ is a pixel in the RGB image, $I_d (p)$  is the average gradient magnitude across color channels in direction $d$. The coefficient $\lambda_c$ controls edge sensitivity.

\xhdr{Overall loss.} The final training objective combines all losses:
\begin{equation}
\gL_{\mathtt{cross{\text -}crw}} + \gL_{\mathtt{intra{\text -}crw}} + \lambda_{s}\gL_{\mathtt{smooth}}
\end{equation}
\section{Experiments}

\begin{table*}[ht]
\centering 
{
\setlength{\tabcolsep}{4pt}

\small

\caption{\small {\small \bf RGB-Depth and RGB-Thermal Matching}. We compare our method with supervised and several unsupervised baselines on NYU-Depth V2, Thermal-IM and KAIST datasets. Our method performs best when computing cross-modality matching, significantly outperforming the baselines. D: Depth, T: Thermal, SSL: Self-Supervised Learning} %

\vspace{-1mm}

\begin{tabularx}{1\linewidth}{X c cccc cccc cc} 

\toprule
\multirow{4}{*}{\textbf{Method}} & \multirow{4}{*}{\textbf{SSL}} & \multicolumn{4}{c}{\textbf{NYU-Depth}, $<\delta^x_\textrm{avg}$~$\uparrow$} & \multicolumn{4}{c}{\textbf{Thermal-IM}, $<\delta^x_\textrm{avg}$~$\uparrow$} & \multicolumn{2}{c}{\textbf{KAIST}, $<\delta^x_\textrm{avg}$~$\uparrow$} \\
\cmidrule(lr){3-6} \cmidrule(lr){7-10} \cmidrule(lr){11-12} 
& & \multicolumn{2}{c}{\textbf{Intra-modal}} & \multicolumn{2}{c}{\textbf{Cross-modal}} & \multicolumn{2}{c}{\textbf{Intra-modal}} & \multicolumn{2}{c}{\textbf{Cross-modal}} & \multicolumn{2}{c}{\textbf{Cross-modal}} \\
\cmidrule(lr){3-4} \cmidrule(lr){5-6} \cmidrule(lr){7-8} \cmidrule(lr){9-10} \cmidrule(lr){11-12} 
& & {\footnotesize RGB $\veryshortarrow$ RGB} & {\footnotesize D $\veryshortarrow$ D} & {\footnotesize RGB $\veryshortarrow$ D} & {\footnotesize D $\veryshortarrow$ RGB} & {\footnotesize RGB $\veryshortarrow$ RGB} & {\footnotesize T $\veryshortarrow$ T} & {\footnotesize RGB $\veryshortarrow$ T} & {\footnotesize T $\veryshortarrow$ RGB} & {\footnotesize RGB $\veryshortarrow$ T} & {\footnotesize T $\veryshortarrow$ RGB} \\
\midrule
RAFT~\cite{teed2020raft} &  & \bf{91.5} & 59.2 & 7.9 & 1.3 & 81.7 & 53.3 & 5.6 & 0.9 & 29.2 & 7.4 \\

GMFlow~\cite{xu2022gmflow} & & 79.7 & 58.2 & 12.7 & 12.5 & \bf{82.2} & \bf{53.4} & 3.8 & 2.6 & 23.1 & 22.3 \\

Arar et al.~\cite{arar2020unsupervised} & \cmark & - & - & 1.3 & 0.8 & - & - & 2.3 & 1.9 & 2.1 & 4.7 \\
{\footnotesize CycleGAN$_{\text{RGB}\scriptveryshortarrow\text{D/T}}$ + GMFlow} & \cmark & - & - & 8.5 & 7.8 & - & - & 7.9 & 7.1 & 8.3 & 8.2 \\ 
{\footnotesize CycleGAN$_{\text{D/T}\scriptveryshortarrow\text{RGB}}$ + GMFlow} & \cmark & - & - & 16.2 & 16.6 & - & - & 6.1 & 5.8 & 6.8 & 7.4 \\
ARFlow~\cite{liu2020learning} & \cmark & 76.1 & 53.5 & 7.5 & 7.4 & 82.1 & 53.3 & 12.5 & 13.2 & 31.0 & 30.4 \\
ARFlow (Retrained) & \cmark & - & - & 9.3 & 8.1 & - & - & 13.4 & 13.1 & 29.1 & 27.7 \\
DIFT~\cite{tang2023emergent} & \cmark & 40.6 & 18.5 & 3.3 & 4.3 & 68.2 & 50.3 & 17.5 & 18.2 & 7.1 & 8.8 \\
SD-DINO~\cite{zhang2023tale} & \cmark & 25.2 & 13.9 & 7.8 & 6.4 & 44.5 & 49.9 & 29.3 & 34.9 & 19.6 & 20.6 \\
Ours & \cmark & 80.1 & \bf{62.3} & \bf{33.5} & \bf{34.3} & 81.6 & 53.1 & \bf{41.8} & \bf{47.9} & \bf{35.2} & \bf{34.1} \\ 
\bottomrule
\label{tab:nyu_depth}
\end{tabularx}
}

\vspace{-6mm}

\end{table*}

Our method estimates pixel-level correspondences between pairs of images across different modalities. We evaluate its performance on both geometric and semantic correspondence tasks. For geometric matching, we consider two settings: RGB-Depth and RGB-Thermal. For RGB-Depth, we use the NYU Depth V2 dataset~\cite{silberman2012indoor} for both training and evaluation. For RGB-Thermal, we train and evaluate on two datasets: the indoor Thermal-IM dataset~\cite{tang2023happened} and the outdoor KAIST dataset~\cite{hwang2015multispectral}. Notably, our method does not require spatially or temporally aligned image pairs during training. We sample frames at different time steps, allowing for scene changes and motion, while assuming partial scene overlap. For semantic correspondence, we consider two tasks. The first is photo-sketch alignment, using the PSC6K dataset~\cite{lu2023learning}, which includes annotated keypoints that match real-world photos with corresponding human-drawn sketches. The second task is cross-style image matching, where the goal is to match corresponding points across stylistic variants (e.g., photorealistic, anime, watercolor) of the same scene. We construct this benchmark using a text-to-image generation model conditioned on consistent prompts and varying style modifiers. For all benchmarks, evaluation sets are created by manually annotating keypoints across modalities or, where applicable, by propagating tracked points from the RGB domain using a point-tracking algorithm.

\subsection{Geometric Correspondence}

\xhdr{RGB-Depth Training.} We train our model on the NYU Depth V2 dataset~\cite{silberman2012indoor}, which contains approximately 400K unlabeled RGB-D frames. Training is performed in three stages: 1) intra-modality CRW applied to RGB-RGB and Depth-Depth pairs, 2) additionally cross-modality CRW applied to RGB-Depth and Depth-RGB pairs, 3) adding the smoothness loss to encourage spatial coherence. During training, we randomly sample two frames ($I^{m_1}_{t}$, $I^{m_2}_{t+k}$) from a scene, separated by a few random timesteps, drawn from both modalities. We then apply different random resized crops to forward ($I^{m_1}_{t}$, $I^{m_2}_{t+k}$) and backward images (${I^{m_1}_{t, \mathtt{aug}}}$)~\cite{shrivastava2024gmrw} and train for cycle-consistency losses.

\xhdr{RGB-Thermal Training.} We use the Thermal-IM and KAIST datasets for RGB-Thermal training. Thermal-IM consists of 783 video sequences with RGB and thermal modalities, though they are spatially unaligned. KAIST contains 320 video sequences with spatially aligned RGB and thermal views. As with RGB-Depth training, we randomly sample two frames per scene from both modalities and follow the same three-stage training pipeline with intra- and cross-modality CRW losses.

\xhdr{RGB-Depth Evaluation.} Since RGB and depth frames in NYU are spatially aligned, we use PIP++~\cite{zheng2023point}, a point tracking method, to generate ground-truth correspondences. We create 10-frame video clips and track points across them, retaining those visible in all frames. These tracks are used to evaluate RGB-RGB, Depth-Depth, and RGB-Depth correspondence. Our final dataset includes 250 video clips with an average of 688 annotated tracks per clip.

\xhdr{RGB-Thermal Evaluation.} Thermal-IM videos are not spatially aligned, so we manually annotate 100 RGB-Thermal frame pairs across 5 timesteps, with 10 keypoints each, following the protocol from TAP-Vid~\cite{doersch2022tap}, resulting in 1,000 evaluation points. For KAIST, where RGB and thermal frames are aligned, we use a strategy similar to NYU Depth, using CoTracker~\cite{karaev23cotracker} to generate correspondences, as it performs better than other trackers on KAIST videos.

\xhdr{Evaluation Metrics.} We adopt the positional accuracy metric from TAP-Vid~\cite{doersch2022tap}, denoted as  $<\delta^x_\textrm{avg}$. This metric reports the fraction of visible keypoints that are predicted within a pixel distance threshold from ground truth, averaged over thresholds ${1, 2, 4, 8, 16}$.

\subsection{Semantic Correspondence}
\label{sec:semantic_corr}
\xhdr{Photo-Sketch Matching.} We use the PSC6K dataset~\cite{lu2023learning} for both training and evaluation. The dataset contains 6,250 annotated photo-sketch pairs, with 8 keypoints per pair in the evaluation set, and over 130K photo-sketch pairs in the training set. Performance is measured using Percentage of Correct Keypoints (PCK) at thresholds $\alpha = 0.05$ and $0.1$, denoted as PCK-5 and PCK-10.

\xhdr{Cross-style Image Matching.} We introduce a new benchmark to evaluate correspondence across different visual styles of the same scene. We generate images using the Flux text-to-image model~\cite{flux2024}, conditioned on a base prompt and a style modifier (added to the prompt). We consider 10 distinct styles: anime, dark, light, photorealistic, pixel art, watercolor, comic book, neon, pastel, and sci-fi (examples in Figure~\ref{fig:cross_style_matching}). Prompts are generated using BLIP-2\cite{li2023blip} by captioning images from the ImageNet dataset~\cite{russakovsky2015imagenet}, and are then used to generate stylistic variants via Flux.

We generate 10K images per style, resulting in 100K training samples. For evaluation, we sample 50 unseen prompts and generate 10 images per prompt---one in each style---yielding 500 test images. We manually annotate corresponding keypoints across styles for each prompt. All possible pairs of styles are compared within each prompt, resulting in 2,250 total image pair comparisons in the test set.

\section{Results}

We train our model for RGB-Depth, RGB-Thermal, photo-sketch and cross-style matching tasks and present both quantitative and qualitative results alongside comparisons with baseline methods.

\subsection{Geometric correspondence}
Table~\ref{tab:nyu_depth} compares our method against a range of supervised and unsupervised baselines for RGB-Depth and RGB-Thermal matching, evaluated in the direct setting where query frames are matched directly to target frames without chaining~\cite{shrivastava2024gmrw}. Among the supervised baselines, we include RAFT~\cite{teed2020raft} and GMFlow~\cite{xu2022gmflow}, state-of-the-art optical flow models trained on RGB images, applied to thermal and depth data by treating them as RGB inputs. For unsupervised baselines, we evaluate Arar et al.\cite{arar2020unsupervised}, a CycleGAN\cite{CycleGAN2017}-based cross-modal registration method, which struggles to generalize due to the difficulty of translating between modalities without paired supervision. To address this, we introduce a variant (CycleGAN+GMFlow) that first translates depth or thermal inputs to RGB using CycleGAN, followed by matching in the RGB domain using GMFlow. We also test the reverse direction---translating RGB to depth or thermal---before applying GMFlow.  We further evaluate ARFlow~\cite{liu2020learning}, an unsupervised optical flow model trained with photometric losses, and include a retrained version adapted to our setting with modality-specific encoders. Finally, we compare against recent diffusion-based correspondence methods, including DIFT~\cite{tang2023emergent}, which leverages Stable Diffusion features, and SD-DINO~\cite{zhang2023tale}, which combines Stable Diffusion and DINOv2 features for robust geometric and semantic matching. Across both RGB-Depth and RGB-Thermal tasks, our method significantly outperforms all baselines.

\begin{figure}
    \centering
    \includegraphics[width=\linewidth]{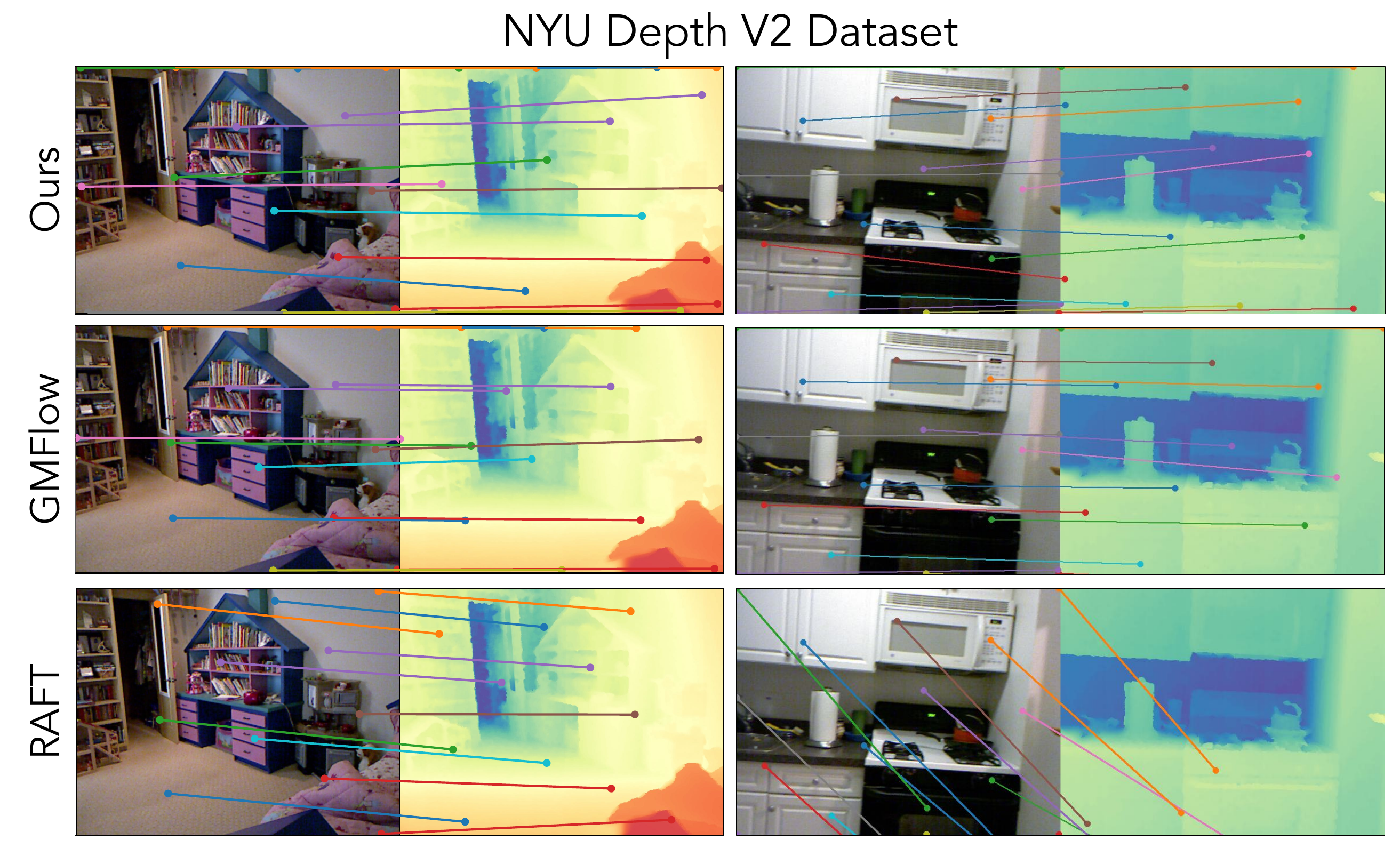}
    \vspace{-5mm}
    \caption{\small {\small \bf RGB-Depth Matching.} We show qualitative comparisons on the NYU Depth V2 dataset. RAFT and GMFlow struggle to establish accurate correspondences in the depth domain, while our method successfully matches keypoints across RGB and depth images. The points shown are randomly sampled from the dataset annotations.}
    
    \label{fig:rgb_depth_qual}
    \vspace{-6mm}
\end{figure}

\begin{figure*}[ht]
    \centering
    \includegraphics[width=\textwidth]{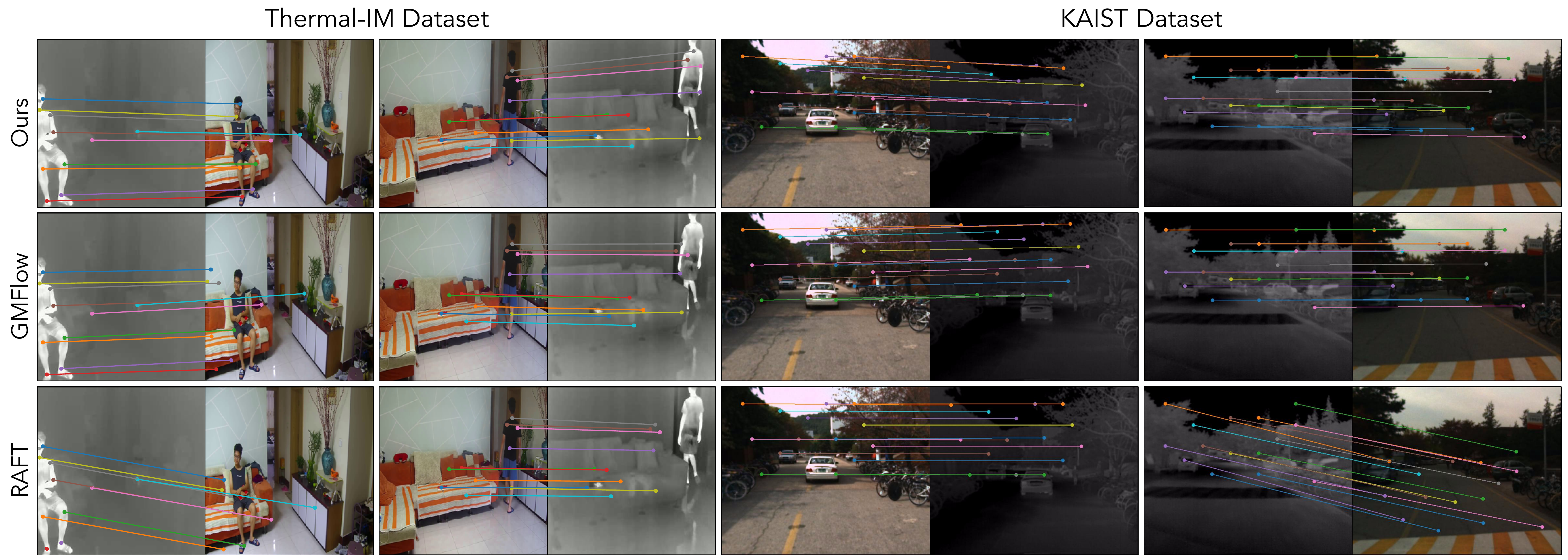}
    \vspace{-5mm}
    \caption{\small{\small \bf RGB-Thermal Matching.} Qualitative comparisons on the Thermal-IM (left) and KAIST (right) datasets. Our method accurately tracks keypoints across RGB and thermal images, even in the presence of motion. In contrast, RAFT and GMFlow often fail to produce meaningful correspondences, frequently matching points to the same spatial location in the other modality. Here, we show all 10 annotated points from the dataset.}
    \label{fig:rgb_thermal_qual}
    \vspace{-2mm}
\end{figure*}

\xhdr{Qualitative results.} We present geometric matching qualitative results showing the effectiveness of our method in cross-modal matching. As shown in Figures~\ref{fig:rgb_depth_qual} and~\ref{fig:rgb_thermal_qual}, our method consistently produces accurate and meaningful correspondences across RGB-Thermal and RGB-Depth pairs, significantly outperforming RAFT and GMFlow, which often yield noisy or incorrect matches.

\xhdr{Model Ablations.} Table~\ref{tab:model_ablations} presents ablations on the NYU-Depth and Thermal-IM datasets. Training with only intra-modal losses yields good performance within the same modality but fails on cross-modal tasks. In contrast, using only cross-modal losses leads to unstable training and poor convergence. Pretraining with intra-modal losses followed by joint intra- and cross-modal training significantly improves performance across both settings. Adding a smoothness loss further enhances cross-modal accuracy, highlighting its value as a regularization strategy.

\subsection{Semantic correspondence}
\begin{figure*}[h]
    \centering
    \includegraphics[width=\textwidth]{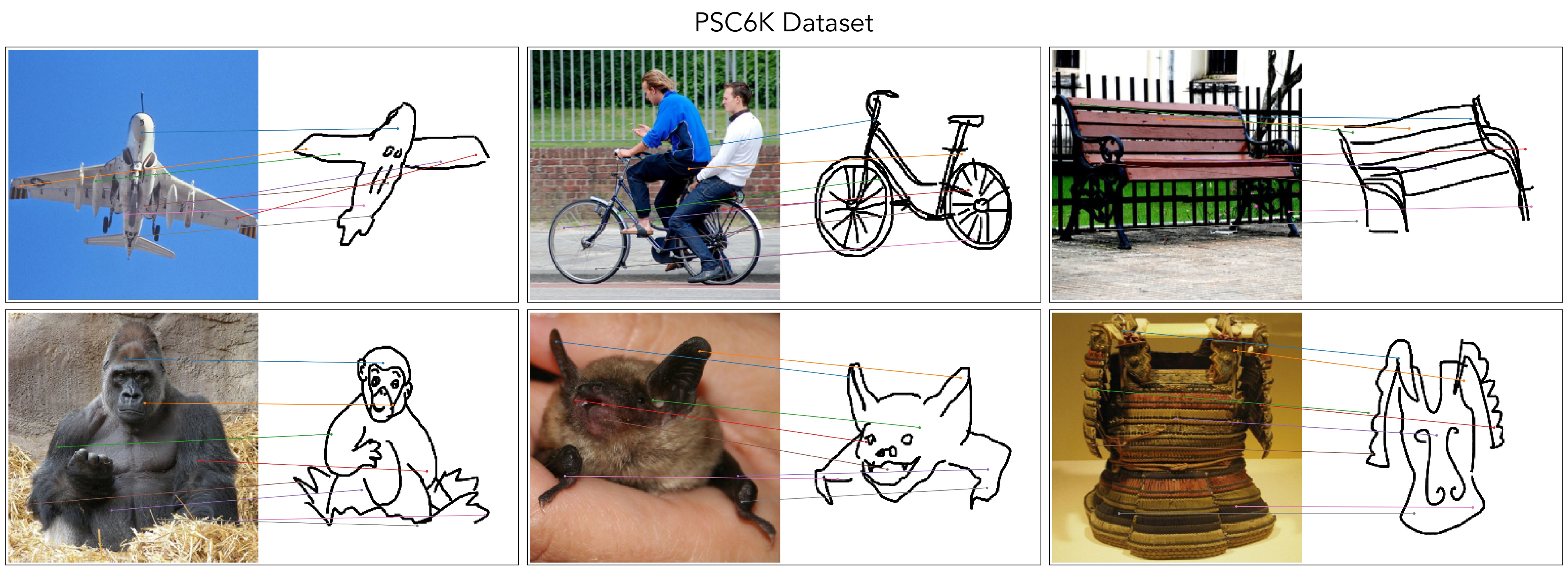}
    \vspace{-5mm}
    \caption{\small{\small \bf Photo-Sketch Matching.} Qualitative results of our method on the PSC6K dataset~\cite{lu2023learning}. For each image pair, all annotated points from the dataset are used as query keypoints.}
    \label{fig:photo_sketch_fig}
    \vspace{3mm}
\end{figure*}

\xhdr{Photo-Sketch Matching.} In Table~\ref{tab:photo_sketch}, we compare our method on PSC6K with several baselines from~\cite{lu2023learning}, including DINOv2+NN~\cite{oquab2023dinov2}, which performs nearest-neighbor search on DINO features, and SD-DINO~\cite{zhang2023tale}. When using a CNN encoder, our method performs poorly due to the lack of semantic priors. Replacing it with DINOv2 features and fine-tuning the encoder leads to a significant performance boost, highlighting the importance of DINOv2's semantic pretraining. Our final model, with DINOv2 and full training, performs best among our variants and is competitive with state-of-the-art methods. Qualitative results are shown in Figure~\ref{fig:photo_sketch_fig}.

\begin{figure*}[h]
    \centering
    \includegraphics[width=\textwidth]{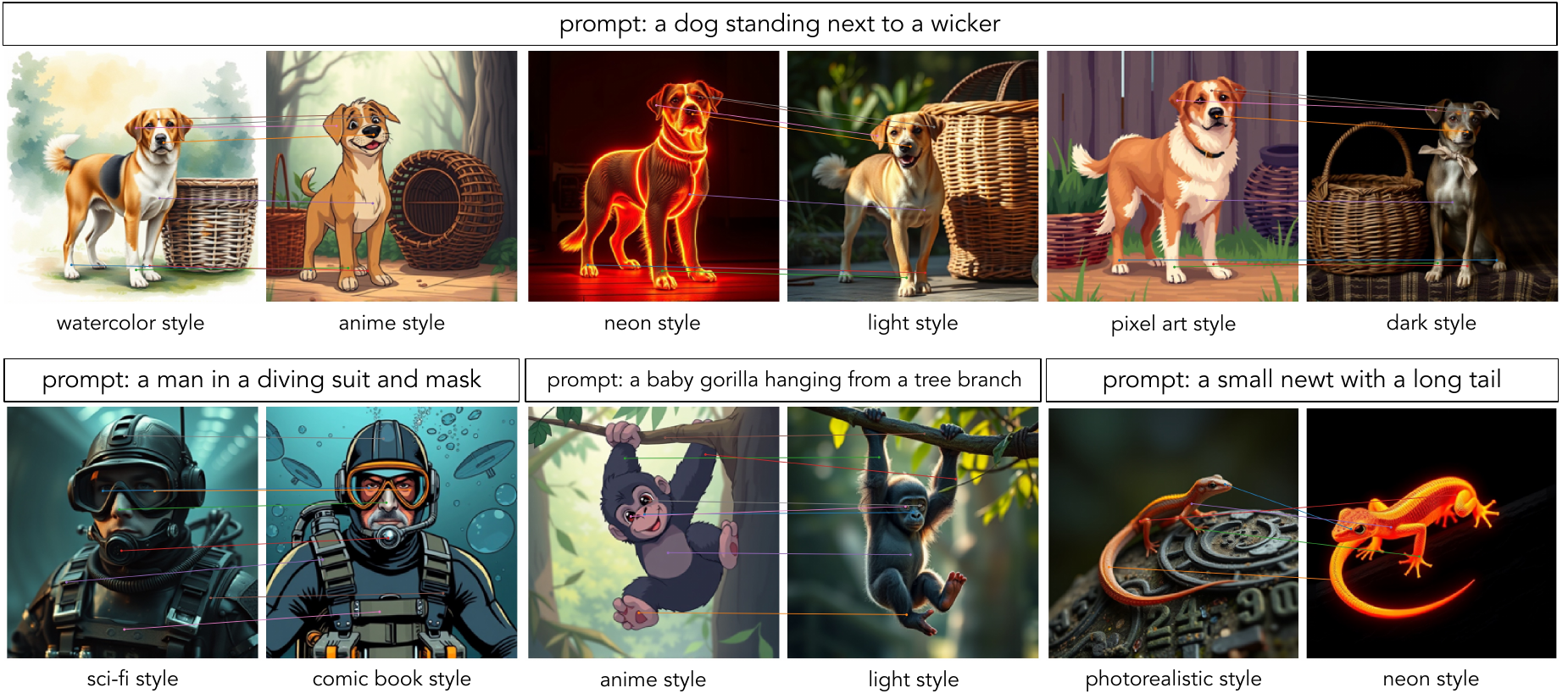}
    \vspace{-5mm}
    \caption{\small{\small \bf Cross-style Image Matching.} We show qualitative results of our method on matching images across different styles. The images are generated using the Flux~\cite{flux2024} model, with the prompts and styles used for generation shown above.}
    \label{fig:cross_style_matching}
    \vspace{-2.5mm}
\end{figure*}

\xhdr{Cross-style Image Matching.} We evaluate our model (using a DINOv2 encoder) alongside baselines including DINOv2+NN, DIFT, and SD-DINO in Table~\ref{tab:cross_style}. We also compare with GeoAwareSC~\cite{Zhang_2024_CVPR}, a state-of-the-art supervised semantic correspondence method. Our model outperforms all baselines, including the supervised approach, albeit by a small margin. The performance gap between our method and DINOv2+NN demonstrates the effectiveness of our learned correspondence framework over simple feature similarity using pretrained DINO features. Qualitative results are presented in Figure~\ref{fig:cross_style_matching}.

\begin{table*}[]
\centering 
\setlength{\tabcolsep}{5pt}

\caption{\small{\small \bf Model Ablations.} We evaluate the impact of different loss functions and show that pretraining with intra-modal losses followed by fine-tuning with all losses achieves the best performance for our method. X denotes D (Depth) or T (Thermal), based on the dataset.}

\vspace{-2mm}

\begin{tabular}{ccccc cccc cc}
\toprule
\multicolumn{5}{c}{\multirow{3}{*}{\textbf{Losses}}} & \multicolumn{4}{c}{\textbf{NYU Depth}, $<\delta^x_\textrm{avg}$~$\uparrow$} & \multicolumn{2}{c}{\textbf{Thermal-IM}, $<\delta^x_\textrm{avg}$~$\uparrow$} \\
\cmidrule(lr){6-9} \cmidrule(lr){10-11} 
& & & & & \multicolumn{2}{c}{\textbf{Intra-modal}} & \multicolumn{2}{c}{\textbf{Cross-modal}} & \multicolumn{2}{c}{\textbf{Cross-modal}} \\
\cmidrule(lr){1-5} \cmidrule(lr){6-7} \cmidrule(lr){8-9} \cmidrule(lr){10-11}
$\gL_{\mathtt{intra{\text -}crw}}^{\texttt{RGB} \veryshortarrow \texttt{RGB}}$ &
$\gL_{\mathtt{intra{\text -}crw}}^{\texttt{X} \veryshortarrow \texttt{X}}$ &
$\gL_{\mathtt{inter{\text -}crw}}^{\texttt{RGB} \veryshortarrow \texttt{X}}$ &
$\gL_{\mathtt{inter{\text -}crw}}^{\texttt{X} \veryshortarrow \texttt{RGB}}$ &
$\gL_{\mathtt{smooth}}$ &
{\footnotesize RGB $\veryshortarrow$ RGB} & {\footnotesize D $\veryshortarrow$ D} & {\footnotesize RGB $\veryshortarrow$ D} & {\footnotesize D $\veryshortarrow$ RGB}  & {\footnotesize RGB $\veryshortarrow$ T} & {\footnotesize T $\veryshortarrow$ RGB}\\
\midrule
\cmark & \cmark & - & - & - & 78.8 & 49.2 & 2.5 & 2.2 & 4.9 & 6.2  \\
- & - & \cmark & \cmark & - & 18.5 & 4.4 & 5.6 & 4.5 & 6.2 & 8.3 \\
\cmark & \cmark & \cmark & \cmark & -  & 80.4 & 61.1& 19.1 & 21.1 & 30.2 & 38.5 \\
\cmark & \cmark & \cmark & \cmark & \cmark & 80.1 & 62.3 & 33.5 & 34.3 & 41.8 & 47.9 \\
\bottomrule
\label{tab:model_ablations}
\end{tabular}

\vspace{-3.5mm}

\end{table*}

\begin{table}[tbp]
\centering 
\small

\caption{\small {\small \bf Photo-Sketch Matching}. Qualitative comparison on the PSC6K dataset~\cite{lu2023learning}, demonstrating the competitive performance of our method against several baselines.} %
\vspace{-2mm}
\def\arraystretch{1}
\setlength{\tabcolsep}{4pt}
\begin{tabularx}{1\linewidth}{X ccc} 
\toprule
\multirow{2}{*}{\textbf{Method}} & \textbf{Trained} & \multirow{2}{*}{\textbf{PCK-5}} & \multirow{2}{*}{\textbf{PCK-10}} \\
&  \textbf{on PSC6K}  & & \\
\midrule
CNNGeo~\cite{rocco2018end} & & 27.59 & 57.71 \\
CNNGeo~\cite{rocco2018end} & \cmark & 19.19 & 42.57 \\
DINOv2 + NN~\cite{oquab2023dinov2} & & 11.48 & 31.66 \\
SD-DINO~\cite{zhang2023tale} & & 33.10 & 70.50 \\
WeakAlign~\cite{rocco2018end} & & 35.65 & 68.76 \\
WeakAlign~\cite{rocco2018end} & \cmark & 43.55 & 78.60 \\
NC-Net~\cite{rocco2018neighbourhood} & & 40.60 & 63.50 \\
DCCNet~\cite{huang2019dynamic} & & 42.43 & 66.53 \\
PMD~\cite{li2021probabilistic} & & 35.77 & 71.24 \\
WarpC-Net~\cite{truong2021warp} & & 48.79 & 71.43 \\
WarpC-Net~\cite{truong2021warp} & \cmark & 56.78 & 79.70 \\
PSCNet~\cite{lu2023learning} & \cmark & {\bf 57.92} & {\bf 84.72} \\

\cdashline{1-4}\noalign{\vskip 0.7ex}
Ours (CNN + Stage 1, 2, 3) & \cmark & 26.22 & 60.89 \\

Ours (DINO + Stage 2, 3) & \cmark & 50.66 & 80.70 \\
Ours (DINO + Stage 1, 2, 3) & \cmark & 53.61 & 82.20\\
\bottomrule
\end{tabularx}

\vspace{-3mm}
\label{tab:photo_sketch}
\end{table}

\begin{table}[tbp]
\centering 
\small
\caption{\small{\small \bf Cross-style Image Matching (Sec.~\ref{sec:semantic_corr}).} Quantitative comparisons on the cross-style image matching task. We compare our method against a few self-supervised and one supervised baseline, and show that it outperforms both.}
\vspace{-2mm}
\def\arraystretch{1.1}
\setlength{\tabcolsep}{4pt}
\begin{tabularx}{1\linewidth}{X ccc} 
\toprule
\textbf{Method} & \textbf{SSL} & \textbf{PCK-5} & \textbf{PCK-10} \\
\midrule
DINOv2 + NN~\cite{oquab2023dinov2} & \cmark & 23.13 & 56.52 \\
DIFT~\cite{tang2023emergent} & \cmark & 39.81 & 57.97 \\
SD-DINO~\cite{zhang2023tale} & \cmark & 60.14 & 80.72 \\
GeoAwareSC~\cite{Zhang_2024_CVPR} & & 68.30 & 79.67 \\
Ours (DINO) & \cmark & {\bf 69.26} & {\bf 84.19} \\

\bottomrule
\end{tabularx}

\vspace{-4mm}
\label{tab:cross_style}
\end{table}

\section{Discussion}
\label{sec:discussion}
We have presented a modality-agnostic method for learning pixel-level correspondences between images taken using different visual sensors. Our method extends the contrastive random walk to perform cross-modal matching. A random walker transitions from patches in one modality to another, and then back, using transition probabilities that are produced by a global matching transformer~\cite{shrivastava2024gmrw,arar2020unsupervised}. We avoid local minima in this process by including intra-modal random walks between frames of the same modality, as well as a spatial smoothness constraint.  We evaluate our method on diverse tasks---including RGB-to-depth, RGB-to-thermal, photo-to-sketch, and cross-style image matching---and find that it significantly outperforms existing methods on geometric correspondence and is competitive with state-of-the-art methods on semantic matching. We see our work as opening new directions in multimodal matching, making it significantly easier to match images without hand-crafted measures of visual similarity. Our approach can be trained entirely using multimodal video data, which can be acquired at scale. We also see our work as experimentally demonstrating the generality of self-supervised matching methods, which can be applied in scenarios that would be difficult to simulate or acquire labeled data for.

\xhdr{Limitations.} We have demonstrated our model for four domains, RGB-depth, RGB-thermal, photo-sketch, cross-style matching. While the model successfully handles these cases, it is possible that other modalities would present additional challenges, especially if they lacked other distinctive visual structures that may be used in matching, such as occluding contours (which are visible in all 3 current modalities, RGB, Thermal, Depth). In cross-style image matching, we observe occasional failures, such as confusion between left and right limbs in animals (see Figure~\ref{fig:cross_style_matching}, top-right). While our method places few assumptions on the underlying signal (e.g., no hand-crafted photo-consistency assumption), both of our datasets include RGB images due to their ubiquity.

\newpage

\xhdr{Acknowledgements.} We thank Xuanchen Lu, Adam Harley, Qianqian Wang, Daniel Geng, Ziyang Chen, Jeongsoo Park, Yiming Dou  and the reviewers for the valuable discussion and feedback. This work was supported by Toyota Research Institute and Cisco Systems.

{
    \small
    \bibliographystyle{cvpr2025template/ieeenat_fullname}
    \bibliography{main}
}

\clearpage

\renewcommand{\thesection}{A.\arabic{section}}
\setcounter{section}{0}

\section{Implementation details}

Here, we present the model architecture in detail and the hyperparameters used during training and evaluation.
\subsection{Model Architecture}
The input to our model is a pair of images of modalities $m_1$, $m_2$ $(I^{m_1}_1, I^{m_2}_2)$ of size $(H, W)$ and the output is an affinity matrix $A$ of size $(H, W, H, W)$ where $A[i,j,k,l]$ represents the probability of pixel $(i, j)$ in $I^{m_1}_1$ transitioning to pixel $(k, l)$ in $I^{m_2}_2$.   

\xhdr{CNN Encoder.} Our feature encoder is similar to GMRW~\cite{shrivastava2024gmrw}, we extract the features at 1/4 scale. We train a different encoder for each modality. The CNN architecture is as follows:

\begin{table}[ht]
\centering
\small
\begin{tabular}{l p{0.2cm} c p{0.2cm} c}
\toprule
\textbf{Layer} & & \textbf{Output} & & \textbf{Details} \\ 
\midrule
\texttt{input} & & $H, W, 3$ &  \\
\texttt{conv1} & & $H/2, W/2, 64$ & & kernel $7\times7$, stride 2 \\
\texttt{res1} & & $H/2, W/2, 96$ & & kernel $3\times3$, stride 1 \\
\texttt{res2} & & $H/2, W/2, 96$ & & kernel $3\times3$, stride 1 \\
\texttt{res3} & & $H/4, W/4, 128$ & &kernel $3\times3$, stride 2 \\
\texttt{res4} & & $H/4, W/4, 128$ & & kernel $3\times3$, stride 1 \\
\texttt{res5} & & $H/4, W/4, 128$ & &kernel $3\times3$, stride 1 \\
\texttt{res6} & & $H/4, W/4, 128$ & & kernel $3\times3$, stride 1 \\

\bottomrule
\end{tabular}
\vspace{0.05in}
\caption{CNN Architecture}
\label{tab:cnn_architecture}
\end{table}

\xhdr{Transformer.} In our transformer model, we stack 6 layers of transformer blocks. Each block consists of a self-attention, cross-attention and feed-forward network. The transformer feature dimension is 128, and the feed-forward network, consisting of 2 linear layers, expands the dimension by $4\times$. For efficiency, we use shifted local window attention~\cite{liu2021swin} where the attention window is $1/16$th the size of the input image.

\subsection{Training Details}
We train our network in Pytorch~\cite{paszke2019pytorch}, using the AdamW~\cite{loshchilov2017decoupled} optimizer with the constant learning rate of $1.6 \times 10^{-4}$. We train on 4 A40 GPUs with an effective batch size 24. For RGB-to-Depth matching, we train the model in three stages: Stage 1 for 50K iterations, Stage 2 for 100K iterations, and Stage 3 for an additional 20K iterations. This full schedule takes approximately 7 days. For RGB-to-Thermal matching, we train Stage 1 for 30K iterations, Stage 2 for 100K iterations, and Stage 3 for another 20K iterations. For Photo-to-Sketch matching, we initialize the visual encoder with a pretrained DINOv2 model, and train for 12K, 10K, and 28K iterations in Stages 1, 2, and 3 respectively with a learning rate of $1.6 \times 10^{-7}$. During Stage 3, we linearly increase the weight of the smoothness loss $\lambda_{s}$ over the first 10K iterations.

\xhdr{Data augmentation.}
For data augmentation, we perform different resize-crop transformations $T^f$ and $T^b$ on the forward and backward cycle in the contrastive random walk. To implement this, we use the \texttt{kornia}~\cite{eriba2019kornia} library. We use it to apply the same transformation ($T^f$/$T^b$) to the entire forward/backward cycle and also use $T^f$ and $T^b$ to compute $T_f^b$ and warp the label i.e. $T_f^b(I)$. We use \texttt{RandomResizedCrop} with range of size ratio of cropped area set to $(0.08, 1.0)$ and the range of aspect ratio of cropped area set to $(0.7, 1.3)$. We set these augmentation hyperparameters by following GMRW~\cite{shrivastava2024gmrw}.

\begin{figure}[h]
    \centering
    \includegraphics[width=\linewidth]{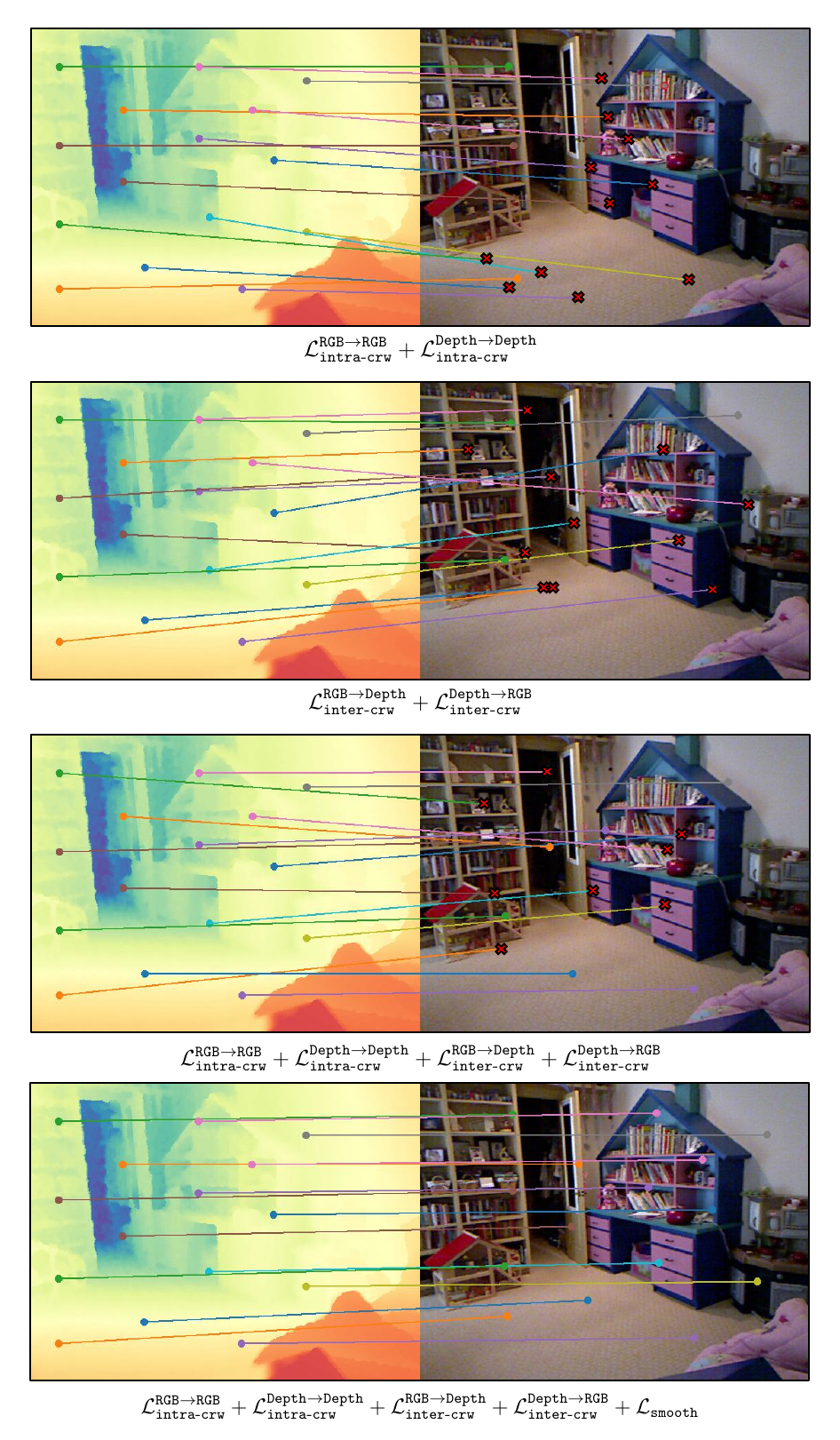}\vspace{-2mm}
    \caption{\footnotesize {\bf Qualitative examples for RGB-Depth matching with different losses. Zoom in for details.} Points with \redcolor{red cross} show the incorrect correspondences (not within 50 px distance of ground truth).  }
    \label{fig:thermal_im_ablations_qual}
\end{figure}

\begin{figure}
    \centering
    \includegraphics[width=\linewidth]{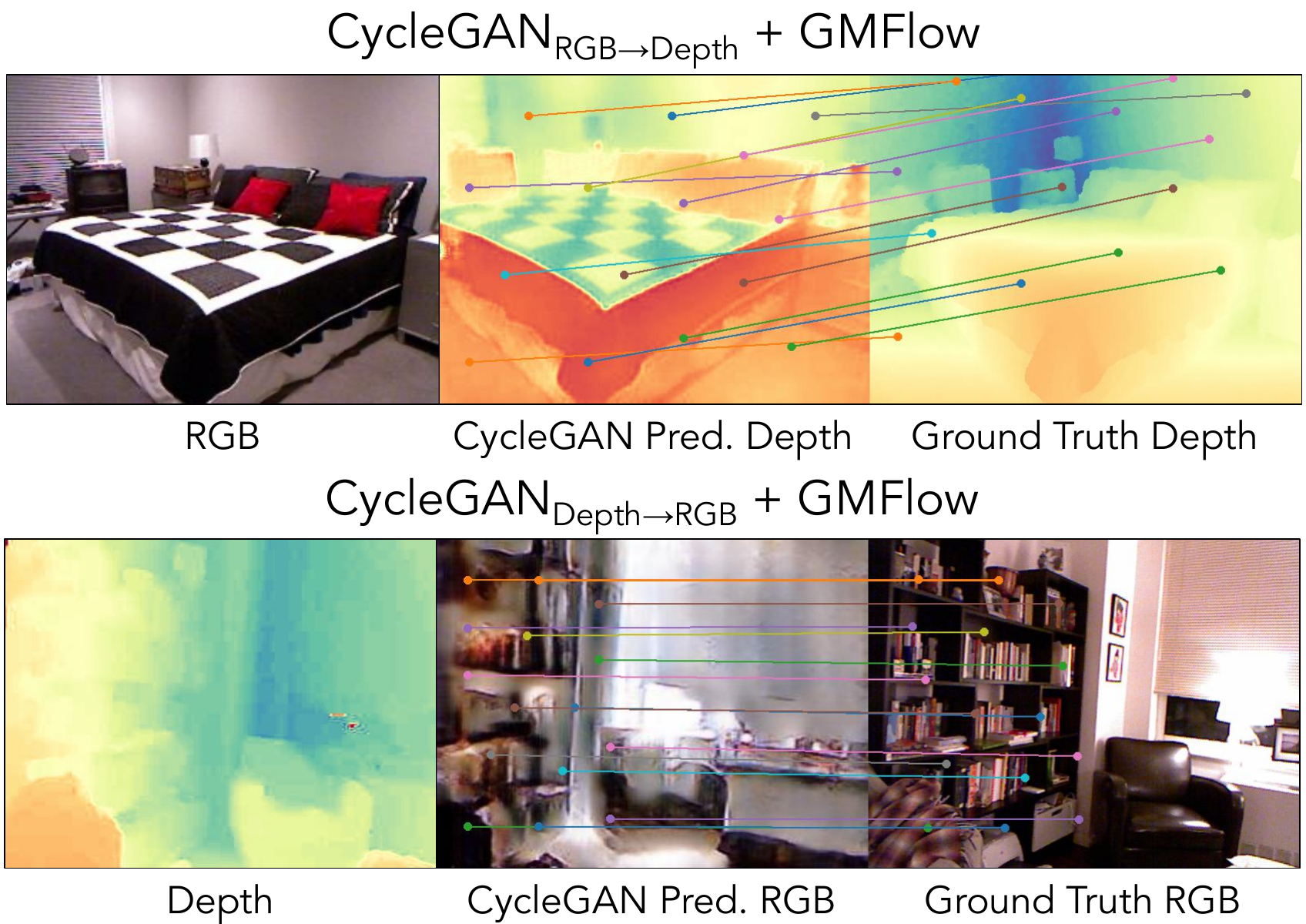}
    \vspace{-6mm}
    \caption{{\bf CycleGAN+GMFlow Baseline.} We show results for a CycleGAN-based baseline that performs image translation from RGB to the depth domain, followed by matching in the depth space. We also present the reverse setup, where images are translated from depth to RGB, and matching is performed in the RGB space. In both cases, the presence of translation artifacts leads to inaccurate matching.}
    \label{fig:cyclegan_qual}
\end{figure}

\subsection{Hyperparameters}
Here are the rest of the hyperparameters used in the model and in training.
\begin{table}[htp]
\centering
{\resizebox{\linewidth}{!}{
\small
\begin{tabular}{l c}
\toprule
\textbf{Hyperparameter} & 
\textbf{Value} \\ 
\midrule

Learning rate schedule  & Constant \\
Learning rate &  $1.6 \times 10^{-4}$ \\
Optimizer & AdamW \\    
Weight Decay & $10^{-4}$ \\
Temperature $\tau$ & $\sqrt{128}$\\
Effective Batch size & 24 \\
Time stride $k$ & Randomly chosen \\
& from [1, 10]\\
Smoothness loss weight $\lambda_s$ & linear increase of $[0, 1]$ over\\
& $[100\text{k}, 120\text{k}]$ steps for RGB-Depth\\

\bottomrule
\end{tabular}
}}
    \caption{\bf Hyperparameters for training the model.}
\end{table}

\begin{figure}
    \centering
    \includegraphics[width=\linewidth]{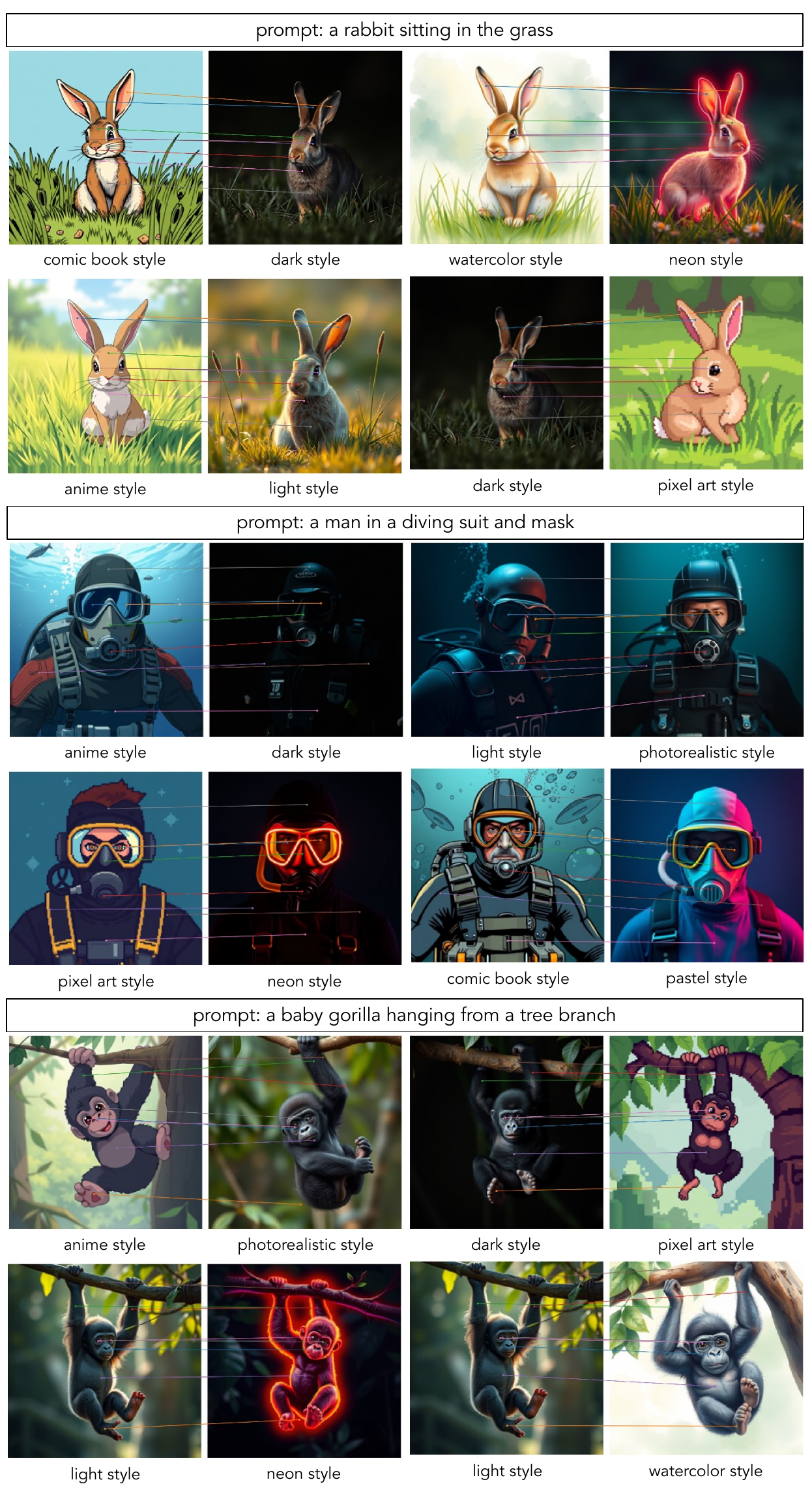}
    \vspace{-5mm}
    \caption{{\bf Cross-style Matching.} We show qualitative results from our model on matching images generated in different styles by a image generation model.}
    \label{fig:suppl_cross_style_matching}
\end{figure}

\section{Dataset details}
\xhdr{NYU Depth Evaluation Dataset.} We leverage the fact that high-quality pseudo ground-truth correspondences can be generated by matching visual frames using off-the-shelf multi-frame tracking methods, in combination with the known calibration between RGB and depth sensors. Specifically, we use PIP++~\cite{zheng2023point} to track points across video clips of length 10. We retain only those tracks that are consistently visible across all frames, reducing the impact of occlusions and tracking failures. We further perform manual visual inspection to ensure the tracks are of reasonable quality and do not drift from their original trajectories. Since RGB and depth frames are spatially aligned via sensor calibration, these tracks can serve as ground truth for RGB-RGB, Depth-Depth, and RGB-Depth matching. The dataset comprises 250 video clips, each with 10 frames and an average of 688 annotated tracks per clip.

\xhdr{Thermal-IM Evaluation Datasets.} In the Thermal-IM dataset, RGB and thermal images are not spatially aligned. To establish ground-truth correspondences, we follow a protocol similar to TAP-Vid~\cite{doersch2022tap}, manually annotating 100 RGB-Thermal image pairs across 5 timesteps with 10 keypoints per frame, resulting in a total of 1000 evaluation points. For the KAIST dataset, where RGB and thermal images are aligned, we adopt a strategy similar to that used for the NYU Depth dataset. Ground-truth correspondences for RGB-RGB, RGB-Thermal, and Thermal-Thermal matching are obtained using CoTracker~\cite{karaev23cotracker} on RGB sequences for videos of length 5.

\section{Qualitative examples}
\begin{figure*}
    \centering
    \includegraphics[width=\linewidth]{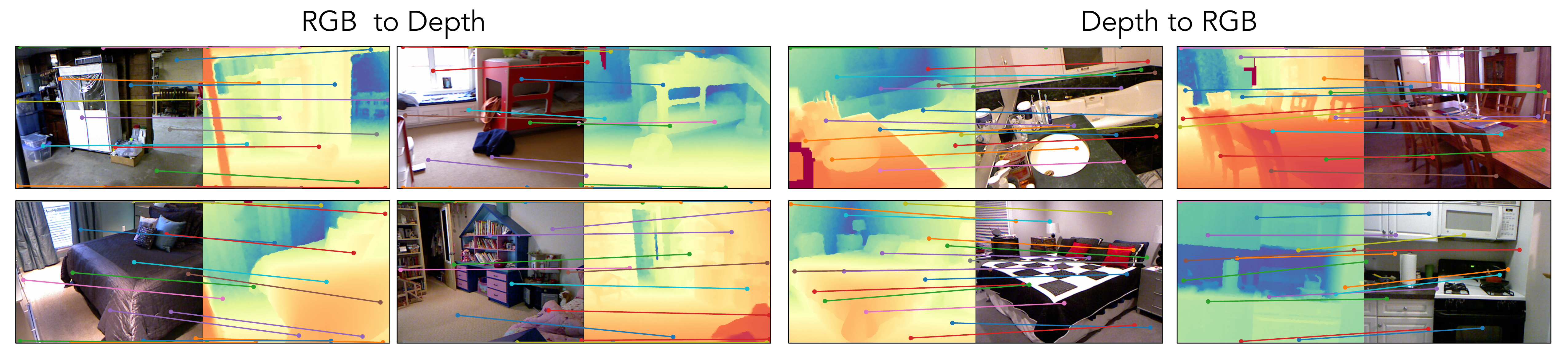}
    \vspace{-5mm}
    \caption{{\bf RGB-Depth Matching.} We show qualitative results from our model on NYU-Depth V2 dataset~\cite{silberman2012indoor}.}
    \label{fig:suppl_rgb_depth_qual}
\end{figure*}

\begin{figure*}
    \centering
    \includegraphics[width=\linewidth]{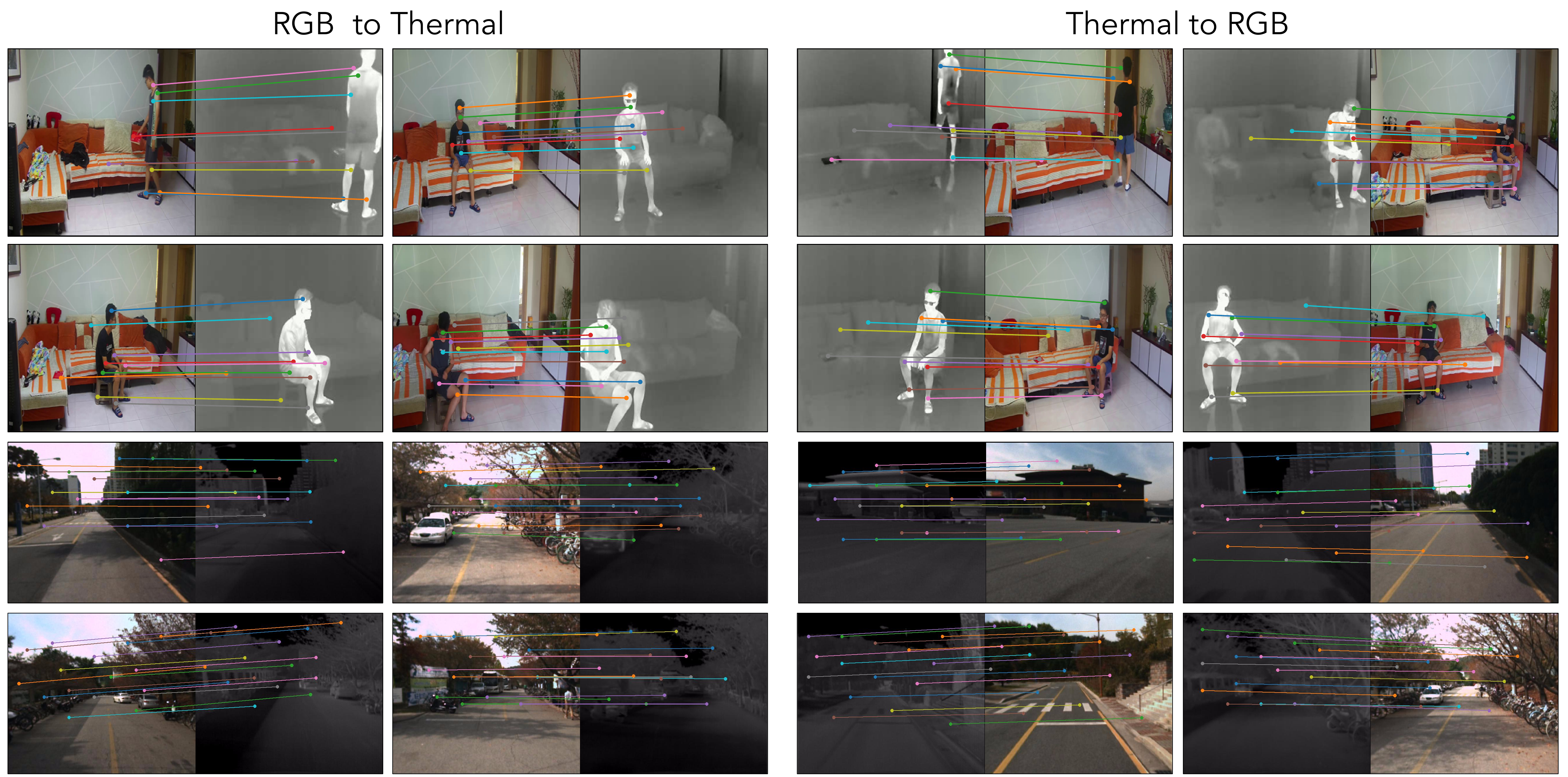}
    \vspace{-5mm}
    \caption{{\bf Thermal-IM Matching.} We show qualitative results from our model on Thermal-IM~\cite{tang2023happened} and KAIST datasets~\cite{hwang2015multispectral}.}
    \label{fig:suppl_rgb_thermal_qual}
    \vspace{-7mm}
\end{figure*}

\begin{figure*}
    \centering
    \includegraphics[width=\linewidth]{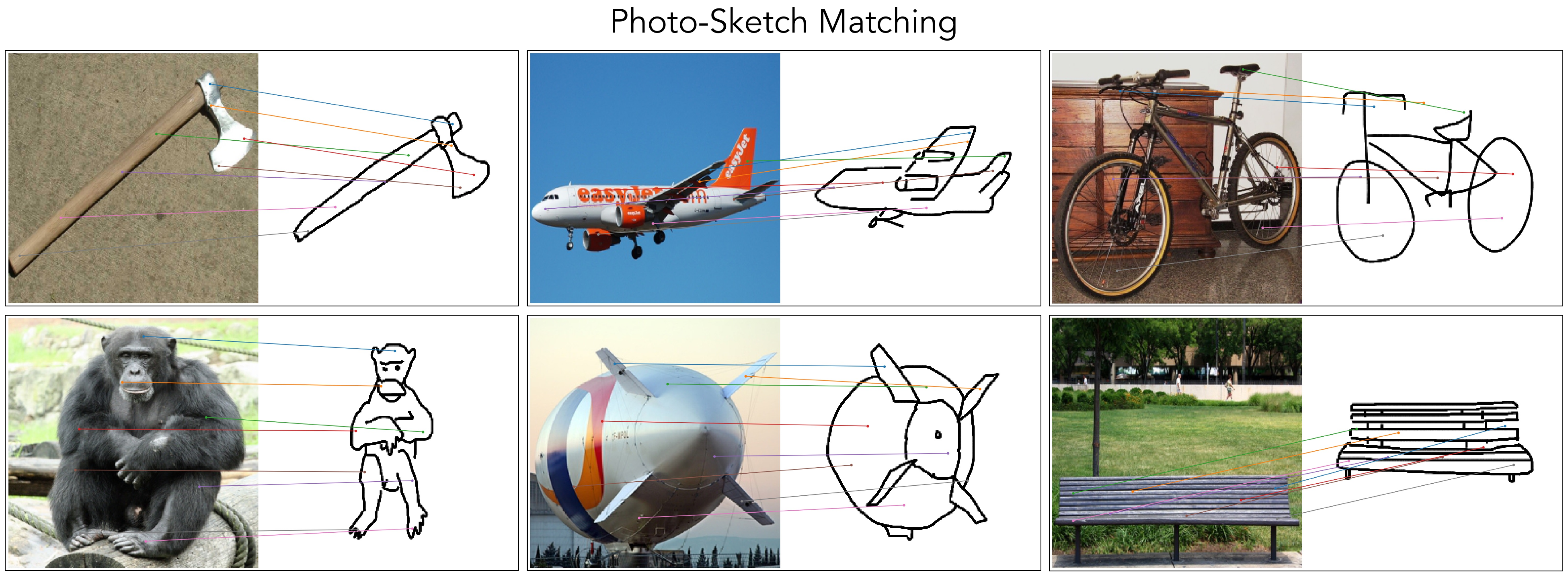}
    \vspace{-5mm}
    \caption{{\bf Photo-Sketch Matching.} We show qualitative results from our model on PSC6K dataset~\cite{lu2023learning}.}
    \label{fig:suppl_photo_sketch_qual}
\end{figure*}

We present additional qualitative results for our model trained on RGB-Depth matching in Figure~\ref{fig:suppl_rgb_depth_qual}, and RGB-Thermal matching in Figure~\ref{fig:suppl_rgb_thermal_qual}. For semantic correspondence tasks, we include more examples for Photo-to-Sketch matching in Figure~\ref{fig:suppl_photo_sketch_qual}, and Cross-style matching in Figure~\ref{fig:suppl_cross_style_matching}.

We also provide qualitative results for the CycleGAN+GMFlow baseline from Table~\ref{tab:nyu_depth}, shown in Figure~\ref{fig:cyclegan_qual}. In this setup, a CycleGAN is trained to translate RGB images to depth, and the pretrained GMFlow model is applied to estimate correspondences between the generated and target depth images. Notably, the CycleGAN often assigns inconsistent depth values to pixels with different colors, even when they belong to the same physical depth plane, resulting in inaccurate matches. We also present results from the reverse setup, where depth images are translated to RGB, and matching is performed in the RGB space. In this case, the generated RGB images often contain significant artifacts and fail to accurately reconstruct the original appearance, degrading correspondence quality.

\end{document}